\documentclass[10pt, a4paper]{article}

\usepackage{authblk}
\usepackage{graphicx}
\usepackage{hyperref}
\usepackage[ruled,vlined]{algorithm2e}

\begin{document}

\title{Bootstrap Bias Corrected Cross Validation applied to Super Learning}

\author[1]{Krzysztof Mnich 
 [0000-0002-6226-981X]}
\author[2]{Agnieszka Kitlas Goli\'nska 
 [0000-0001-8737-765X]}
\author[2]{Aneta Polewko-Klim 
 [0000-0003-1987-7374]}
\author[1,2,3]{Witold R. Rudnicki 
 [0000-0002-7928-4944]}
\affil[1]{Computational Centre, University of Bialystok}
\affil[2]{Institute of Informatics, University of Bialystok}
\affil[3]{Interdisciplinary Centre for Mathematical and Computational Modelling,        University of Warsaw}
\date{}

\maketitle

\begin{abstract}
Super learner algorithm can be applied to combine results of multiple base learners to improve quality of predictions. The default method for verification of super learner results is by nested cross validation.

It has been proposed by Tsamardinos et al., that nested cross validation can be replaced by resampling for tuning hyper-parameters of the learning algorithms. We apply this idea to verification of super learner and compare with other verification methods, including nested cross validation.

Tests were performed on artificial data sets of diverse size and on seven 
real, biomedical data sets. 
The resampling method, called Bootstrap Bias Correction, proved to be a reasonably precise and very cost-efficient alternative for nested cross validation.

\end{abstract}
\setcounter{tocdepth}{1}

\section{Introduction}

Numerous machine learning algorithms with roots in various areas of computer science and related fields have been developed for solving different classes of problems.  
They include linear models, support vector machines, decision trees, ensemble algorithms like boosting or random forests, neural networks etc \cite{fernandez2014we}. 
There are also many feature selection techniques used to prepare the input data for the predictive algorithm. 
Different methods can extract and utilise different parts of the information contained in the data set under scrutiny.
It has been shown that in many cases an ensemble machine learning model, which combines several different predictions, can outperform the component learners. 
This can be realised in the form of wisdom of crowds \cite{marbach2012wisdom} or in a more systematic way as a Super Learner  \cite{superlearner}. 
The wisdom of crowds is one of the principles underlying the design of DREAM Challenges, where multiple team contribute diverse algorithms and methodologies for solving complex biomedical problems \cite{saez2016crowdsourcing}. 
Super learning was proposed by van der Laan et al. and implemented as SuperLearner R language package. 
It utilises cross validation to estimate the performance of the component algorithms and dependencies between them. 
One may notice that wisdom of crowds can be formally cast as a special example of super learning. 
The goal of the current study is to examine  several methods for estimate the performance of the super learner algorithm. 
In particular it explores methods for minimising bias with minimal computational effort. 

\subsection{Super learning -- basic notions and algorithms} 
The input of any machine learning algorithm is a set of $N$ observations, usually assumed to be independent and identically distributed, of the response variable $y_j$ and a set of $p$ explanatory variables $X_j=\{x_{jm}\}$, $m=1,\ldots,p$. 
A predictive algorithm $f$ (that includes also the feature selection algorithm and the set of hyper-parameters) can be trained using a data set $D=\{y_j, X_j\}$ to produce a model $M$. 
Then, the model can be applied to another set of variables $X'$ to obtain a vector of predictions $\Psi=M(X')$.

$K$-fold cross validation is an almost unbiased technique to estimate performance of a machine learning algorithm when applied to unseen data. 
Algorithm \ref{alg_cv} allows to compute a vector of $N$ predictions for all the observations in the data set. 
The predictions can be compared with the original decision variable $Y$ to estimate the performance of the learning algorithm.

\begin{algorithm}[!tb]
\label{alg_cv}
\SetAlgoLined
\DontPrintSemicolon
\SetKwInOut{Input}{input}\SetKwInOut{Output}{output}
\Input{learning method $f$,\\ data set $D=\{y_j,X_j\}$, $j=1,\ldots,N$}\;
\Output{$N$ predictions for the response variable $\psi_j$, $j=1,\ldots, N$}\;
 split randomly the data set into $k$ almost equally-sized folds $F_i$\;
 \ForEach{$F_i$}{
  define a training set as $D_i=D\setminus F_i$\;
  learn a predictive model using the training set $M_i\leftarrow f(D_i)$\;
  apply the model to the remaining subset $\Psi_i\leftarrow M_i(F_i)$\;
  collect the predictions $\Psi_i$ \;
 }
 \caption{$k$-fold cross validation ${\bf CV}(f, D)$}
\end{algorithm}

In super learning, we use multiple machine learning algorithms $f_l$, $l-1, \ldots, L$. 
The cross validation procedure leads to $L$ vectors of predictions $\Psi_l$. 
The idea is to treat them as new explanatory variables and apply some machine learning algorithm to build a second-order predictive model (see Algorithm \ref{alg_sl}). 
The new features are expected to be strongly and linearly connected with the response variable, so linear models with non-negative weights seem to be appropriate super learning algorithms. Indeed they are default methods in SuperLearner R package. 
Note, however, that any method that selects a weighted subset of the elementary learning algorithms can be formally considered as and example of super learning. 
This includes also selection of the best-performing algorithm, which is a common application of cross validation, and can be considered as a special case of super learning \cite{superlearner}. 
Other examples can be unweighted mean of $k$ best performing algorithms, as in the wisdom of crowds, or even a simple mean of all learning algorithms used.

\begin{algorithm}[htbp]
\label{alg_sl}
\SetAlgoLined
\DontPrintSemicolon
\SetKwInOut{Input}{input}\SetKwInOut{Output}{output}
\Input{$L$ learning methods $f_l$,\\ combining method $f_c$,\\ data set $D=\{Y,X\}$,\\
 new data set $D'=\{Y',X'\}$}\;
\Output{ensemble predictive model $M$,\\ predictions for the new response variable $\Psi'$}\;
 compute $L$ vectors of cross validated predictions $\Psi_l\leftarrow {\bf CV}(f_l,D)$\;
 build a second-order data set $D_c\leftarrow \{Y, \Psi_l\}$\;
 apply the method $f_c$ to the second-order data set $M_c\leftarrow f_c(D_c)$\;
 \BlankLine
 build $L$ predictive models using the entire data set $M_l\leftarrow f_l(D)$\;
 apply the models to the new data $\Psi'_l\leftarrow M_l(D')$\;
 compute the ensemble predictions $\Psi'\leftarrow M_c(\Psi'_l)$\;
 \KwRet{$M=\{M_l, M_c\}$, $\Psi'$}\;
 \caption{Super learning ${\bf SL}(f_l, f_c, D, D')$}
\end{algorithm}

\subsection{Performance estimation for super learning -- nested cross validation}

Super learning, as every machine learning method, is sensitive to overfitting. 
Therefore, an unbiased estimate of the performance of ensemble models is necessary. 
The obvious and most reliable method to obtain it is an external cross validation. 
The entire procedure is called nested cross validation, as it contains two levels of CV loop (see Algorithm \ref{alg_nested}).

\begin{algorithm}[!t]
\label{alg_nested}
\SetAlgoLined
\DontPrintSemicolon
\SetKwInOut{Input}{input}\SetKwInOut{Output}{output}
\Input{$L$ learning methods $f_l$,\\ combining method $f_c$,\\ data set $D=\{Y,X\}$}\;
\Output{predictions for the response variable $\Psi$}\;
 split randomly the data set into $k$ almost equally-sized folds $F_i$\;
 \ForEach{$F_i$}{
  define a training set as $D_i=D\setminus F_i$\;
  run a super learning procedure $\{M_i, \Psi_i\}\leftarrow {\bf SL}(f_l, f_c, D_i, F_i)$\;
  collect the predictions $\Psi_i$ \;
 }
 compare $\Psi$ with corresponding values of $Y$
 \caption{Nested cross validation ${\bf NCV}(f_l, f_c, D)$}
\end{algorithm}

Nested cross validation is implemented in SuperLearner R package as a default method of performance estimation. 
The authors of Super learner algorithm recommend to use 10-fold internal cross validation  \cite{superlearner}. 
The external CV should also be at least 10-fold, to avoid a meaningful reduction of the sample size. 
Even if one restricts himself to a single loop of the external CV, the entire procedure requires all the learners to be run 110 times. 
Although the routine is easy to parallelise, the computational complexity is very large.

\subsection{Bootstrap bias correction for super learning}

An alternative approach to verification of complex machine learning algorithms was proposed by Tsamardinos et al. \cite{tsamardinos}. 
It is called the Bootstrap Bias Correction and bases on Efron's bootstrapping technique \cite{bootstrap}. 
The method was originally proposed to estimate the bias caused by a choice of the optimal set of hyper-parameters for a predictive model. 
However, as it has been mentioned, this task can be considered as a special case of super learning. 
What is more, the procedure is general and does not involve any actions specific to selection of hyper-parameters. 
Thus, the Tsamardinos' method can be easily generalised for any kind of super learning.

The idea is shown in Algorithm \ref{alg_bbc}. 
The cross validated predictions for all the algorithms are computed only once. 
Then, the combining models are computed many times for samples that are drawn with replacement from the predictions. 
The predictions of each combination model are then tested on it's out-of-bag observations. 

\begin{algorithm}[t]
\label{alg_bbc}
\SetAlgoLined
\DontPrintSemicolon
\SetKwInOut{Input}{input}\SetKwInOut{Output}{output}
\Input{$L$ learning methods $f_l$,\\ combining method $f_c$,\\ data set $D=\{Y,X\}$}\;
\Output{set of pairs of the response variable and predictions $\{y_j,\Psi_j\}$}\;
 \ForEach{$f_l$}{
  compute the cross validated predictions $\Psi_l\leftarrow {\bf CV}(f_l,D)$\; 
  build a second-order data set $D_c\leftarrow \{Y, \Psi_l\}$\;
 }
 \For{$b\leftarrow 1$ \KwTo $B$}{
  draw a random sample $D_b$ of the prediction set $D_c$ with replacement\; 
  define the out-of-bag set $D_{\setminus b}\leftarrow D_c\setminus D_b$\;
  apply the method $f_c$ to the data set $D_b$ $M_b\leftarrow f_c(D_b)$\;
  compute predictions for the out-of-bag set  $\Psi_b\leftarrow M_b(D_{\setminus b})$\;
  compare the predictions $\Psi_b$ and with the corresponding values of $Y$\;
 }
 \caption{Bootstrap bias corrected super learning ${\bf BBCSL}(f_l, f_c, D)$}
\end{algorithm}

In this method, all base predictions come from the same cross-validation run, hence the entire sample is somewhat co-dependent. Therefore learning on the outcome will be overfitted. On the other hand, the training set contains duplicates what introduces additional noise. The effective data set size is smaller than the sample. This effect should decrease the overfitting and cancel the bias of the performance estimate.

The major advantage of BBC method is its small computational complexity. The elementary predictions are computed only once, multiple runs are required only for the relatively simple combining procedure. 

In the current study, we apply bootstrap bias corrected super learning algorithm to synthetic and real-world data sets  and compare the results with other verification methods, including the nested cross validation.

\section{Materials and Methods}

The tests were focused on binary classification tasks. 
The area under ROC curve (AUC) was used as a quality measure of predictions, since it does not depend on the class balance in the data set.  
The methods, however, can be applied also for multi-class classification or regression tasks, with different quality metrics. 

\subsection{Data}

\subsubsection{Artificial data}
The methodology was developed on the synthetic data set. 
The data set was created as follows:
\begin{itemize}
    \item First, the two vectors of expected values and two covariance matrices were randomly generated. 
    These parameter sets are common for all the observations.
    \item The requested number of instances of the binary decision variable was randomly chosen, with the same probability for both classes.
    \item For each class of the decision variable, the explanatory variables are drawn from a multivariate normal distribution with the corresponding set of parameters.
\end{itemize}

This procedure allows to create an arbitrary big sample with the same statistical properties and verify directly the predictions on the unseen data. 
The parameters were tuned to emulate the strength of linear, quadratic and pairwise interactions as well as the dependencies between variables that may appear in the real-world data sets.

The data sets we had generated contained 5000 explanatory variables and 50, 100, 150, 200 observations. Statistic tests indicate from 2 to 20 relevant variables, depending on the sample size.

\subsubsection{Biomedical data sets}
The tests were performed on seven 
data sets that contain measurements of gene expression and copy number variation for four cancer types. 
These data sets correspond to biological questions with different levels of difficulty. 
These are:
\begin{itemize}
\item data sets obtained from the CAMDA 2017 Neuroblastoma Data Integration Challenge (http://camda.info): 
\begin{itemize}
    \item CNV -- 39 115 array comparative genomic hybridization (aCGH) copy number variation profiles, data set limited to a cohort of 145 patients, 
    \item MA -- 43 349 GE profiles analysed with Agilent 44K microarrays, data set limited to a cohort of 145 patients, 
    \item RNA-seq -- 60 778 RNA-seq GE profiles at gene level, data set limited to a cohort of 145 patients. 
\end{itemize}
The data collection procedures and design of experiments were previously described in the original studies \cite{zhang,theissen,stig,coco,kocak}. 
Data sets are also accessible in Gene Expression Omnibus \cite{GEO}.
The relevant question for these data sets is predicting the final clinical status of the patient using molecular data. 
This is difficult problem. 

\item data sets with The Cancer Genome Atlas database generated by the TCGA Research Network
(https://www.cancer.gov/tcga) 
that contain RNA-sequen\-cing data for various types of cancer \cite{Hammerman_2012,Collisson_2014,Koboldt_2012,Ciriello_2015,Lawrence_2015,Creighton_2013}: 
\begin{itemize}
    \item BRCA RNA -- data set contains 1205 samples and 20223 probes (112 normal and 1093 tumor),
    \item BRCA CNV -- data set contains 1314 samples and 21106 probes (669 normal and 645 tumor),
    \item HNSC -- data set contains 564 samples and 20235 probes (44 normal and 520 tumor),
    \item LUAD -- data set contains 574 samples and 20172 probes (59 normal and 515 tumor).  
\end{itemize}
In this case the relevant question is discerning normal tissue from tumor using data set at hand. 
It is much easier question since both genetic profile and gene expression patterns are highly modified in cancer cells in comparison with normal tissue. 
  \end{itemize}
\subsection{Methods}

For each data set we applied the full protocol of super learning and nested cross validation. We used the default 10-fold setup for both external and internal cross validation and 100 repeats of resampling procedure. For the artificial data sets, we performed the entire routine for 100 different sets drawn from the same distribution. In the case of real data sets, the protocol was repeated 100 times on the same data for different cross validation splits.

\subsubsection{Base machine learning algorithms}
Six popular machine learning algorithms were used as base learners: 
\begin{itemize}
    \item Random Forest \cite{randomforest},
    \item LASSO \cite{lasso},
    \item Support Vector Machine (SVM) \cite{svm},
    \item AdaBoost \cite{adaboost},
    \item Naive Bayesian classifier,
    \item kNN classifier for $k=10$. 
\end{itemize}
All the algorithms are implemented from the standard R packages, with the default parameters. The parameters used are obviously not optimal, but the performance optimisation is not the subject of the current study.

For each algorithm, the input data were prepared by the feature selection algorithm. To this end, we applied Welch $t$-test for each explanatory variable and chose 100 variables with the biggest value of the test statistic 
This procedure is very sensitive to overfitting, so any bias in the verification methods should be clearly visible.

\subsubsection{Methods of super learning}
Six methods of combining various prediction results via super learner approach were tested: 
\begin{itemize}
 \item two default methods implemented in SuperLearner R package:
 \begin{itemize}
    \item{\bf NNLS:} non-negative least squares,
    \item{\bf NNlog:} non-negative logistic regression,
 \end{itemize}
 \item two ``toy example'' methods, that are, however, commonly used:
 \begin{itemize}
    \item{\bf Mean:} average of all the base results (the method does not introduce any overfitting),
    \item{\bf Best 1:} choice of the best-fitted model (this special case is mentioned in the original van der Laan paper and corresponds directly to the original purpose of the bootstrap method by Tsamardinos),
 \end{itemize}
 \item{\bf Best $k$:} average of $k$ best-performing models, where $k$ is optimised on the training set -- usually $k$ is set as 3-4,
  \item{\bf RF:} Random Forest algorithm.
\end{itemize}

\subsubsection{Methods of verification} The estimate of quality was performed using the following methods (from the most biased up to unbiased one):
\begin{itemize}
    \item{\bf Training set:} measure the performance of the combined classifier on the same data that were used to build the combined model (the results are obviously overestimated);
    \item{\bf Independent CV:} compute the results of base learners only once, then verify the combining algorithm in a second, independent cross validation (due to the common information in the training and validation sets the results are also overestimated);
    \item{\bf BBC SL:} apply the bootstrap bias correction method (the method proposed in the current study),
    \item{\bf Nested CV:} apply the nested cross validation (as a gold-standard);
    \item{\bf New data:} directly measure the performance for new data, drawn from the same distribution (the oracle for artificial data sets).
\end{itemize}

\section{Results}

\subsection{Artificial data sets}

\subsubsection{Base learners -- cross validation vs. direct verification}

The way of building the artificial data sets allows for comparison between the cross validation results and the actual results obtained for unseen data. This comparison is shown in Table \ref{tab_singlesynt}. 

\begin{table}[t]
\centering
\caption{The AUC of base classification algorithms for the artificial data. 
Comparison between prediction estimate in 10-fold cross-validation ({\bf 10CV}), prediction of new data for model trained on entire sample ({\bf 100\%}) and prediction of new data for model trained on 90\% of the sample ({\bf 90\%}) (the same training set size as in cross-validation). The uncertainty values of AUC were computed as mean square error (MSE) of average of 100 independent measurements.}
{
\begin{tabular}{l|ccc|ccc}
    & \hspace{0.5em}10CV\hspace{0.5em} & \hspace{0.5em}100\%\hspace{0.5em} & \hspace{0.5em}90\%\hspace{0.5em} & \hspace{0.5em}10CV\hspace{0.5em} & \hspace{0.5em}100\%\hspace{0.5em} & \hspace{0.5em}90\%\hspace{0.5em} \\
\hline
 & \multicolumn{3}{c|}{50 obs., MSE=0.01} & \multicolumn{3}{c}{100 obs., MSE=0.01}\\
\hline
RandomForest & 0.62 & 0.64 & 0.62 & 0.69 & 0.71 & 0.69 \\
LASSO        & 0.62 & 0.63 & 0.62 & 0.68 & 0.70 & 0.69 \\
SVM          & 0.63 & 0.64 & 0.62 & 0.68 & 0.71 & 0.69 \\
AdaBoost     & 0.38 & 0.33 & 0.38 & 0.68 & 0.71 & 0.69 \\
Naive Bayes  & 0.63 & 0.65 & 0.63 & 0.69 & 0.72 & 0.70 \\
kNN          & 0.56 & 0.57 & 0.55 & 0.61 & 0.63 & 0.61 \\
\hline
 & \multicolumn{3}{c|}{150 obs., MSE=0.005} &\multicolumn{3}{c}{200 obs., MSE=0.005}  \\
 \hline
RandomForest & 0.735 & 0.760 & 0.745 & 0.782 & 0.804 & 0.788 \\
LASSO        & 0.744 & 0.772 & 0.747 & 0.817 & 0.853 & 0.821 \\
SVM          & 0.738 & 0.754 & 0.744 & 0.777 & 0.803 & 0.787 \\
AdaBoost     & 0.699 & 0.731 & 0.711 & 0.761 & 0.789 & 0.763 \\
Naive Bayes  & 0.735 & 0.756 & 0.744 & 0.770 & 0.792 & 0.779 \\
kNN          & 0.659 & 0.678 & 0.672 & 0.690 & 0.711 & 0.698 \\
\hline
\end{tabular}}
\label{tab_singlesynt}
\end{table}

As it could be expected, the performance of all the machine learning algorithms improves with the sample size. 
In particular AdaBoost could not cope with 50 observations only.

Interestingly, a considerable negative bias of the cross validated estimations of AUC was observed in comparison with the model trained on the entire sample.  
To correct this difference, additional models were built using the training sets reduced to 90\% of the original sample. 
Quality of these models, agree much better with the cross-validated estimates. 
That means that most of the bias of the cross validation procedure comes from the smaller size of training sets. 
The small remaining bias is most likely due to negative correlations of fluctuations in training and validation sets in cross-validation. 

\subsubsection{Super learning}

The performance of diverse methods of super learning is shown in Table \ref{tab_slsynt}. 
For this particular data set, the super learning technique needs at least 100 observations to outperform the best single result and at least 200 observations to perform better then a simple average of all the base results. 
Random forest proves to be a poor super learning method. 
The non-negative linear models perform the best (surprisingly, the ordinary least squares method was better at classification than the logistic regression, that is specialised for this task). 
However, the simple best-$k$ method performs almost as well. 

The differences between various performance estimates are small, but some regular patterns are noticeable.
As previously, the performance of the models learned on the entire data set and applied to new data is better than measured in the nested cross validation. 
Moreover, the performance measured for models built using the reduced training set is slightly better as well. 
The biased performance estimate on the training set is significantly bigger than other estimates.
For 100 and more observations, BBC SL method leads to the results very close to the nested cross validation. The results of the independent CV are more unstable, often overestimated. 

These results are, however, ambiguous: the bias of cross validation methods due to the smaller size of training sets seems to be stronger than any bias due to overfitting. 
Nevertheless, the proposed BBC SL algorithm proved better than any other simple verification algorithm and close to the ``gold standard'' nested cross validation. 

\begin{table}[tb]
\centering
\caption{Artificial data -- AUC of super learning for 6 diverse methods for three sizes of artificial data sets.}
{
\begin{tabular}{l|c|c|c|c|c|c}
\hline
SL    &\hspace{0.1em}Training\hspace{0.1em} &\hspace{0.1em}Indep.\hspace{0.1em}& \hspace{1.0em}BBC\hspace{1.0em} & \hspace{0.5em}Nested\hspace{0.5em} & \hspace{0.05em}New data\hspace{0.05em} & \hspace{0.05em}New data \\
method&set&CV&SL&CV&100\% & 90\% \\
\hline
\multicolumn{7}{c}{50 observations, MSE=0.01}\\
\hline
NNLS&0.66&0.62&0.62&0.59&0.64&0.62\\
NNlog&0.65&0.62&0.62&0.59&0.59&0.58\\
RF&0.55&0.57&0.56&0.52&0.56&0.57\\
Best $k$&0.67&0.64&0.63&0.61&0.63&0.62\\
Mean&0.63&0.63&0.63&0.63&0.64&0.62\\
Best 1&0.67&0.64&0.63&0.61&0.64&0.63\\
\hline
\multicolumn{7}{c}{100 observations, MSE=0.01}\\
\hline
NNLS&0.71&0.69&0.69&0.68&0.71&0.70\\
NNlog&0.71&0.68&0.69&0.68&0.70&0.69 \\
RF&0.65&0.66&0.64&0.63&0.66&0.64 \\
Best $k$&0.73&0.70&0.69&0.68&0.71&0.70 \\
Mean&0.70&0.70&0.70&0.70&0.72&0.70 \\
Best 1&0.72&0.70&0.68&0.67&0.71&0.69 \\
\hline
\multicolumn{7}{c}{200 observations, MSE=0.005 for Best 1 and Best $k$, 0.004 for others}\\
\hline
NNLS&0.829&0.815&0.810&0.814&0.851&0.827\\
NNlog&0.831&0.816&0.805&0.804&0.854&0.828\\
RF&0.800&0.802&0.790&0.792&0.832&0.808\\
Best $k$&0.830&0.812&0.803&0.799&0.852&0.824\\
Mean&0.808&0.808&0.807&0.808&0.834&0.816\\
Best 1&0.822&0.807&0.797&0.794&0.851&0.816\\
\hline
\end{tabular}}
\label{tab_slsynt}
\end{table}

\subsection{Biomedical data}

The results of super learner procedure obtained for real data are displayed in Tables \ref{tab_real1}, \ref{tab_real2}. 
The direct measurement of performance for unseen data is impossible in this case, hence a nested cross-validation is a reference.  
The estimated standard deviation of the distribution of results is shown for each data set.  
The error of the mean value is smaller, but it is hard to estimate, since the measurements are not mutually independent.
As for artificial data, the performance of Random Forest for combining the base results was very poor, and the non-negative logistic regression performed very close to NNLS method. Thus, both were not shown in the tables.

For all the tested data sets ensemble methods outperform the best single classifier. However, in all the cases the best-performing method was a simple average over all the base results. The linear model and $k$-best proved nearly as good in some cases. This result is obviously not a rule, our artificial data sets give an example for better performance of linear combinations of classifiers.

Another interesting point is the stability of the results. The most repeatable values of AUC are produced by mean of all the classifiers, while choice of the best single one and mean of $k$ best ones are the most unstable. The effect is not visible on the training set, but clearly appears when nested CV or BBC SL verification methods are used (see especially Table \ref{tab_real2}). 

The performance measured using the proposed BBC SL method is almost exactly the same as obtained by nested cross validation. Two simpler methods of the performance estimation report inflated results. 

\begin{figure}[h]
    \centering
    \includegraphics[width=0.99\textwidth]{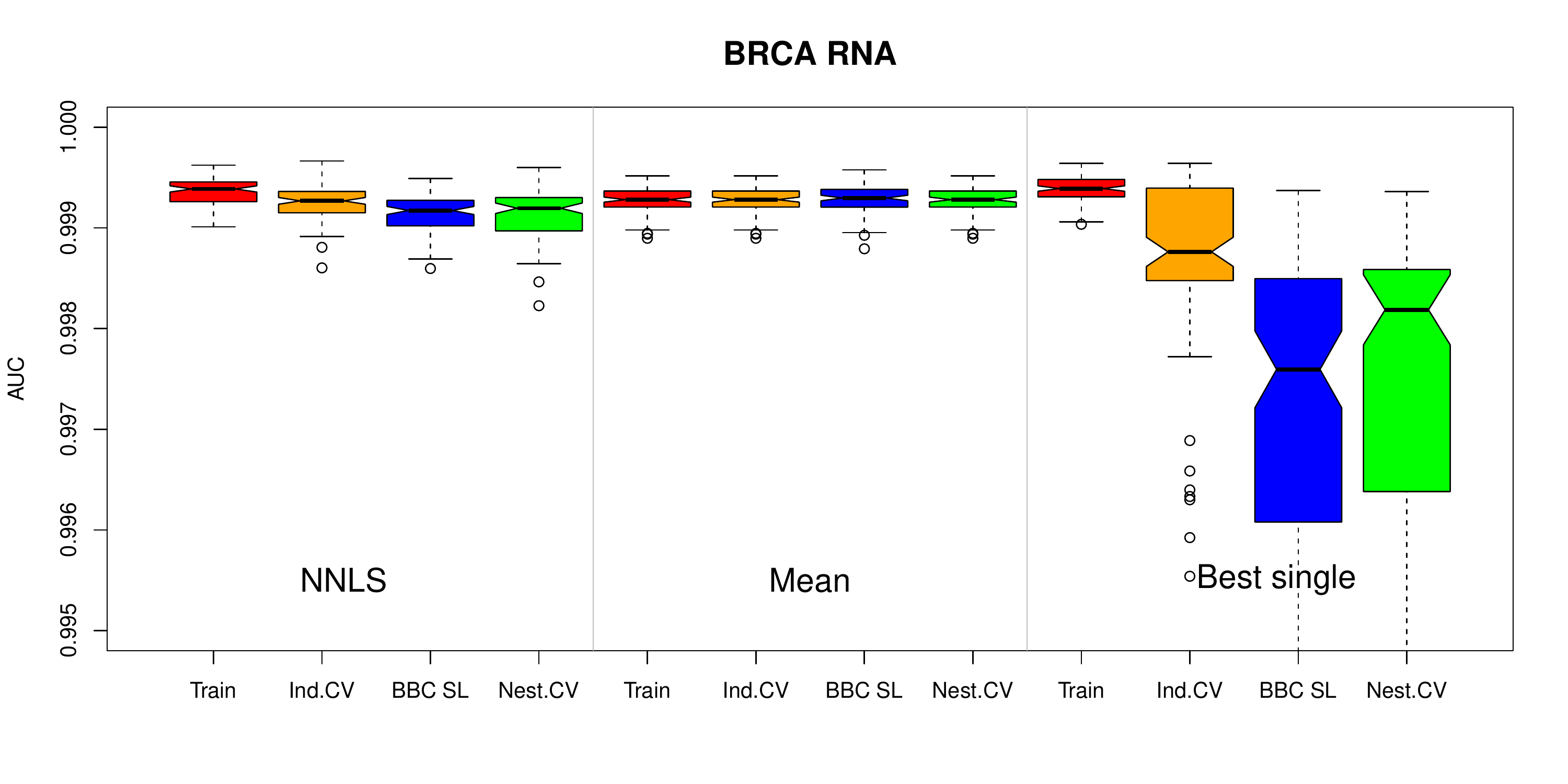}
    \caption{Box plots of AUC for an example real data set. Note the drop of performance and the instability of results for the best single classifier, when unbiased verification methods are applied.}
    \label{fig_boxplot}
\end{figure}

Figure \ref{fig_boxplot}) shows the performance of three combining methods for the BRCA RNA data set in more detailed way.
The ``Best single'' method is simply a common practice of choosing the best-performing classifier to this particular purpose. In a simple cross validation, the result seems to be better than any ensemble model. However, when the external validation is applied, the performance drops significantly and becomes unstable. It turns out, that as simple operation as choice between six classification algorithms is a considerable source of overfitting. Both nested cross validation and the proposed BBC SL method show this clearly.
For this particular data set, the optimal strategy is the average of all the base predictions.

\begin{table}[t]
\centering
\caption{The performance of super learning for 4 methods for three Neuroblastoma data sets. Random Forest and NNlog methods were omitted. Note the mean square error for {\bf Best 1} and {\bf Best $k$} methods, bigger than the typical values.}
\begin{tabular}{l|c|c|c|c}
\hline
SL    &Training&Independent& BBC&Nested \\
method&set&CV&SL&CV \\
\hline
\multicolumn{5}{c}{CNV, typical MSE=0.02}\\
\hline
NNLS    & 0.78 & 0.77 & 0.76 & 0.76 \\
Best $k$& 0.78 & 0.77 & 0.75 &0.76$\pm$0.03 \\
Mean& 0.76 & 0.76 & 0.76 & 0.76 \\
Best 1& 0.78 & 0.77$\pm$0.03&0.74$\pm$0.03&0.75$\pm$0.03\\
\hline
\multicolumn{5}{c}{MA, typical MSE=0.01}\\
\hline
NNLS& 0.89 & 0.87 & 0.87 & 0.88\\
Best $k$& 0.89 &0.88$\pm$0.02&0.87$\pm$0.02&0.88$\pm$0.02\\
Mean& 0.89 & 0.89 & 0.89 & 0.89 \\
Best 1& 0.89 &0.88$\pm$0.02&0.85$\pm$0.02&0.87$\pm$0.02\\
\hline
\multicolumn{5}{c}{RNA, typical MSE=0.01}\\
\hline
NNLS& 0.91 & 0.90 & 0.89 & 0.89\\
Best $k$& 0.91 & 0.90 &0.89$\pm$0.02&0.89$\pm$0.02\\
Mean& 0.89 & 0.89 & 0.89 & 0.89 \\
Best 1& 0.90 &0.90$\pm$0.02&0.88$\pm$0.02&0.88$\pm$0.02\\
\hline
\end{tabular}
\label{tab_real1}
\end{table}

\begin{table}[h]
\centering
\caption{Real data (2) -- the performance of super learning for 4 different methods for TCGA data sets. Note the dependence of the square error on the combining method (the biggest for {\bf Best 1} and {\bf Best $k$}) and on the verification method (smaller on the training set)}
\begin{tabular}{l|c|c|c|c}
\hline
SL    &Training&Independent& BBC&Nested \\
method&set&CV&SL&CV \\
\hline
\multicolumn{5}{c}{BRCA RNA}\\
\hline
NNLS&0.9994$\pm$0.0001&0.9992$\pm$0.0002&0.9991$\pm$0.0002&0.9991$\pm$0.0002 \\
Best $k$&0.9994$\pm$0.0001&0.998$\pm$0.002&0.998$\pm$0.001&0.998$\pm$0.003 \\
Mean&0.9993$\pm$0.0001&0.9993$\pm$0.0001&0.9993$\pm$0.0001&0.9993$\pm$0.0001 \\
Best 1&0.9994$\pm$0.0001&0.998$\pm$0.003&0.997$\pm$0.002&0.996$\pm$0.004 \\
\hline
\multicolumn{5}{c}{BRCA CNV}\\
\hline
NNLS&0.988$\pm$0.001&0.987$\pm$0.001&0.987$\pm$0.001&0.987$\pm$0.001 \\
Best $k$&0.988$\pm$0.001&0.987$\pm$0.001&0.987$\pm$0.001&0.987$\pm$0.001 \\
Mean&0.988$\pm$0.001&0.988$\pm$0.001&0.988$\pm$0.001&0.988$\pm$0.001 \\
Best 1&0.983$\pm$0.001&0.981$\pm$0.003&0.981$\pm$0.002&0.981$\pm$0.002 \\
\hline
\multicolumn{5}{c}{HNSC}\\
\hline
NNLS&0.996$\pm$0.002&0.992$\pm$0.005&0.990$\pm$0.002&0.993$\pm$0.006 \\
Best $k$&0.997$\pm$0.001&0.988$\pm$0.01&0.986$\pm$0.005&0.98$\pm$0.01 \\
Mean&0.995$\pm$0.003&0.995$\pm$0.003&0.995$\pm$0.002&0.995$\pm$0.003 \\
Best 1&0.997$\pm$0.001&0.985$\pm$0.01&0.983$\pm$0.006&0.98$\pm$0.01 \\
\hline
\multicolumn{5}{c}{LUAD}\\
\hline
NNLS&0.9993$\pm$0.0002&0.998$\pm$0.002&0.9978$\pm$0.0007&0.998$\pm$0.002 \\
Best $k$&0.9994$\pm$0.0001&0.998$\pm$0.003&0.997$\pm$0.002&0.997$\pm$0.003 \\
Mean&0.9992$\pm$0.0001&0.9992$\pm$0.0001&0.9991$\pm$0.0002&0.9992$\pm$0.0001 \\
Best 1&0.9993$\pm$0.0001&0.998$\pm$0.004&0.997$\pm$0.002&0.997$\pm$0.003 \\
\hline
\end{tabular}
\label{tab_real2}
\end{table}

\section{Conclusion}

Super learner as proposed in the \cite{superlearner} is computationally demanding approach that relies on multiple cross-validation and application of multiple learning algorithms. 
In the original formulation of the algorithm verification of the quality of the final model involves repeating entire procedure within nested cross-validation, what significantly increases the computational cost of the modelling procedure. 
The current study shows that  the nested cross-validation can be replaced by using resampling protocol, which gives equivalent results. 

In almost all examined cases, the ensemble of learners gives better results than selection of a single best method for prediction. 
Interestingly, the simplest method of combining results of all algorithms, namely the the simple average of predictions of all base learners seems to be a very good choice for obtaining a stable non-overfitted estimate.  
Two simplest methods for assigning weights to base learners, namely a simple linear combination or selection of unweighted average of k-best models have better performance that other methods and should be explored along simple mean of all base methods. 

One should note, that Super learner was applied here to merge results of algorithms that are very good predictors themselves. 
The differences between predictions of these classifiers are concentrated on a handful of difficult cases.  
The simple average works best, because there are too few independent data point to build reliable model for more advanced methods. \bibliographystyle{splncs03_unsrt}
\bibliography{superlearner}

\end{document}